%% file: arxiv_ver.tex
\def\year{2021}\relax
\documentclass[letterpaper]{article} 
\usepackage{aaai21}  
\usepackage{times}  
\usepackage{helvet} 
\usepackage{courier}  
\usepackage[hyphens]{url}  
\usepackage{graphicx} 
\usepackage{amssymb}
\usepackage{multirow}
\usepackage{multicol}
\usepackage{amsmath}
\usepackage{natbib}  
\usepackage{caption} 
\title{Channel-wise Alignment for Adaptive Object Detection}

\author{
	Hang Yang$^{\wr}$~~~~~ Shan Jiang$^{\S}$~~~~~ Xinge Zhu$^{\S}$~~~~~ Mingyang Huang$^{\dag}$\\ Zhiqiang Shen$^{\ddagger}$~~~~~ Chunxiao Liu$^{\dag}$ ~~~~~ Jianping Shi$^{\dag}$ \\
}
\affiliations{
	$^{\wr}$Beihang University ~~~~ $^{\S}$Chinese University of Hong Kong ~~~~ \\$^{\ddagger}$Carnegie Mellon University ~~~~ $^{\dag}$SenseTime Research
}
\begin{document}

	\newcommand{\etal}{\textit{et al}.}
	\newcommand{\zhu}[1]{\textcolor{red}{zhu: #1} }
	\newcommand{\ie}{\textit{i}.\textit{e}.}
	\newcommand{\eg}{\textit{e}.\textit{g}.}

\maketitle

\begin{abstract}

\input{abstract.tex}

\end{abstract}

\section{Introduction}

\input{intro.tex}

\section{Related Work}
\input{related.tex}

\section{Methodology}
\input{method.tex}

\section{Experiments}
\input{exper.tex}

\section{Conclusion}
\input{conclusion.tex}

{\small
\bibliography{egbib}
}
\end{document}

%% file: abstract.tex
Generic object detection has been immensely promoted by the development of deep convolutional neural networks in the past decade. However, in the domain shift circumstance, the changes in weather, illumination, etc., often cause domain gap, and thus performance drops substantially when detecting objects from one domain to another. Existing methods on this task usually draw attention on the high-level alignment based on the whole image or object of interest, which naturally, cannot fully utilize the fine-grained channel information. In this paper, we realize adaptation from a thoroughly different perspective, i.e., channel-wise alignment. Motivated by the finding that each channel focuses on a specific pattern (e.g., on special semantic regions, such as car), we aim to align the distribution of source and target domain on the channel level, which is finer for integration between discrepant domains.
 Our method mainly consists of self channel-wise and cross channel-wise alignment. These two parts explore the inner-relation and cross-relation of attention regions implicitly from the view of channels. Further more, we also propose a RPN domain classifier module to obtain a domain-invariant RPN network. Extensive experiments show that the proposed method performs notably better than
existing methods with about 5\% improvement under various domain-shift settings. Experiments
on different task (e.g. instance segmentation) also demonstrate its good scalability. 

%% file: intro.tex
Deep convolutional neural networks have brought impressive advances across a multitude of tasks in computer vision. At the same time, deep convolutional neural networks need to be supported by the large amount of labeled data~\cite{ma2019trafficpredict,Cordts2016Cityscapes,kitti}, the data should contain all scenarios in the real world so that it can work well for testing (with the similar distributions), but in practice it's hard to collect the data in all situations, also it's costly expensive and impractical. Hence,  
object detectors deploying in the real world still face challenges from the changes in view-points, object appearance, backgrounds, illumination, image quality, etc., which can cause significant domain shift, and thus performance degradation. 



\begin{figure}[t]
   \centering
   \includegraphics[width=0.9\linewidth]{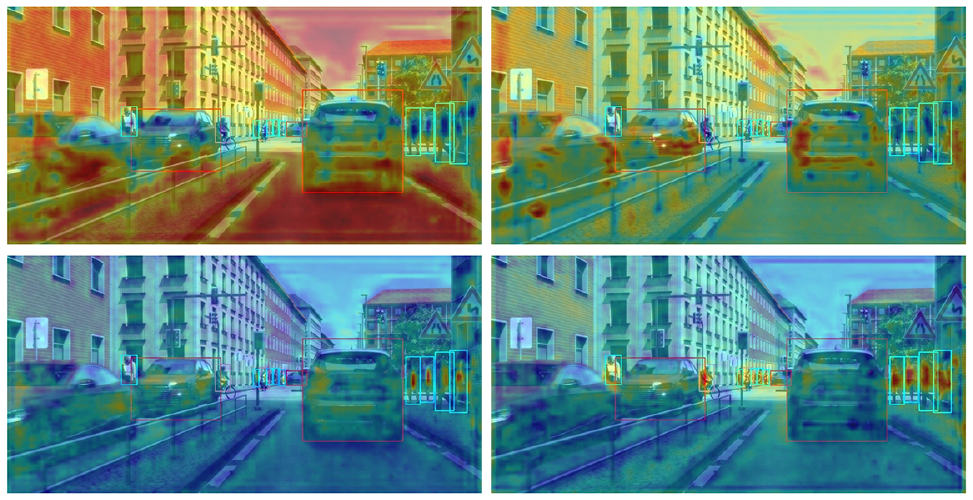} 
   \vspace{-2ex}
   \caption{The visualization of different channel feature maps on the same input image. We can observe clearly that these four visualizations focus on background (upper-left), car (upper-right), person's arms (bottom-left) and person's body (bottom-right) respectively. It indicates that the feature maps from different channels have different weights of contribution to the whole model.}
   \label{fig:attention}
\end{figure}

Traditional approaches for domain adaptation on detection mainly focus on the alignment of semantic features. For instance,~\cite{chen2018domain} aligns the features in the image level and instance level, respectively.~\cite{zhu2019adapting} mines the objects of interest and aims to align these pertinent regions.~\cite{saito2019strong} introduces the weak alignment to similar images and strong alignment to local receptive fields. Orthogonal to those works focusing on high-level feature alignment, we aim to investigate the impact of channel-wise features on domain adaptation for detection. As shown in Fig.~\ref{fig:attention}, we observe that different channels usually focus on the different patterns, such as car or pedestrian. This finding delivers that channel-wise features contain sufficient information covering various patterns, which is favorable for the domain adaptation as source and target domains often have inconsistent semantic features (\ie, different combinations of objects). Motivated by these findings, we propose a new approach to adapt object detectors by channel-wise space alignment.

Specifically, the proposed framework consists of two key components, self channel-wise alignment and cross channel-wise alignment, which respectively address the inner-relation and cross-relation alignment in channel-wise space. Here, self channel-wise alignment module treats each channel as a distinct distribution, and attach the domain classifier to learn the robust features. For the cross channel-wise alignment, we propose a novel decay gram-matrix loss to explore the mutual information between channels. In self channel-wise alignment component, the adversarial training strategy~\cite{ganin2016domain} is applied. We further incorporate an RPN domain classifier to align the features of proposals, which can enhance the robustness to the regions of interest. Overall, the cooperation of these components leads to an adaptation process in the channel-wise space, thus improves the effectiveness.


We conduct experiments on three paired domain-shift datasets, including 
{normal-to-foggy} (Cityscapes~\cite{Cordts2016Cityscapes} to Foggy-Cityscapes~\cite{foggy-cityscapes}),
{synthetic-to-real} (SIM-10k~\cite{sim10k} to Cityscapes), and
{cross-camera} (KITTI~\cite{kitti} to Cityscapes).
On the experimental results, the proposed method yields considerable improvement over existing methods. Furthermore, we also extend our method on instance segmentation task, the remarkable result shows our approach can handle this situation robustly.

Our contributions can be grouped into three aspects. 1) Our studies uncover a new path to the success of detection adaptation, \ie, channel-wise alignment.
2) We develop the channel-wise alignment framework, which consists of self channel-wise alignment and cross channel-wise alignment for performing the inner-relation and cross-relation alignment. 3) We conducted extensive experiments to compare the proposed methods with
others on various settings, where it yields notable performance gains not only
in object detection but also in instance segmentation.

%% file: related.tex

\subsection{Object Detection}
Object detection is one of the most important topic in the field of computer vision. In recent years, most detection networks can be gathered into two categories, that is one-stage and two-stage. 

One-stage detectors do not need Region Proposal Network (RPN) to generate a series of candidate regions. They solve the object classification and box regression problem with only just one single network, such as YOLO \cite{redmon2016you,yolov2,yolov3}, SSD \cite{liu2016ssd} and its variants~\cite{zhu2020ssn,wang2020reconfigurable}.

Two-stage detectors generate a number of proposals using RPN in the first stage, and then fine-tune these proposals using another network in the second stage. Though one-stage detectors reach higher speed than two-stage detectors, many state-of-the-art frameworks still using two-stage detectors, which have higher performance than one-stage detectors, \textit{e.g.} R-CNN \cite{girshick2014rich}, Fast R-CNN \cite{girshick2015fast}, Faster R-CNN \cite{ren2015faster}, and Mask R-CNN \cite{he2017mask}, \textit{etc}.

However, these state-of-the-art object detectors still face the same awkward challenge, which is relying heavily on the annotated datasets in specific scenes. For example, a state-of-the-art detector trained on public dataset like KITTI or Cityscapes, whose performance may be severely degraded when working in real world environment like autonomous driving system due to domain shift. We can solve this problem by collecting better labeled data from target domain, and then fine-tune the detector on these data, yet it is time-consuming and expensive, even the labeled data can't be collected in some cases.

\subsection{Domain Adaptation}
Domain adaptation \cite{ben2010theory,mansour2009domain} has been well-focused in recent years, which is trying to adapt a model trained on one domain to another new domain. At present, a number of domain adaptation works focus on image classification \cite{duan2012domain,duan2011visual,fernando2013unsupervised,gong2012geodesic,gopalan2011domain,kulis2011you} and semantic segmentation \cite{sun2019not,chen2019crdoco,li2019bidirectional,luo2019taking,vu2019advent,zou2018unsupervised,chen2020tensor}. In image classification, some representative methods are effective, including domain transfer multiple kernel learning \cite{duan2012domain,duan2011visual}, asymmetric metric learning \cite{gong2012geodesic}, subspace interpolation \cite{fernando2013unsupervised}, geodesic flow kernel \cite{gopalan2011domain} and subspace alignment \cite{kulis2011you}. In semantic segmentation, Sun \textit{et al.} \cite{sun2019not} propose a cascade-weighted network to measure the similarity between synthetic pixels and real pixels,  \cite{chen2019crdoco,zhu2018penalizing} conduct an adversarial adaptation method based on pixel-wise level or design a loss function to align the domain. \cite{song2020adastereo} proposes three alignment modules to achieve the domain adaptation on stereo matching.

\subsection{Domain Adaptation for Objection Detection}
There have a number of state-of-the-art domain adaptation achievements, yet most of these methods are focus on image classification or semantic segmentation, thus application in the field of object detection is still in early stages~\cite{shen2019scl}. Chen \textit{et al.} \cite{chen2019crdoco} propose to adapt domain gap with both image-level and instance-level feature alignment, furthermore, integrate gradient reverse layer (GRL) \cite{ganin2016domain} into Faster-RCNN framework to reduce the domain gap. Saito \textit{et al.} \cite{saito2019strong} pay attention to low-level and high-level features, and tries to align them. Zhu \textit{et al.} \cite{zhu2019adapting} focus on clustering high relationship regions and aligning these regions.

It is obvious that previous works do not discover the high relationship between channels in feature map. In our work, we design channel-wise alignment method to reduce the gap of source domain and target domain which include two parts, \textit{i.e.} self channel alignment and cross channel alignment. Furthermore, we add an an  adversarial learning part on RPN sub-network, which aims at generating more robust region proposals. Note that our work can be trained in end-to-end fashion, and do not need any target domain annotations.

%% file: method.tex
\begin{figure*}[t]
   \centering
   \includegraphics[width=0.9\linewidth]{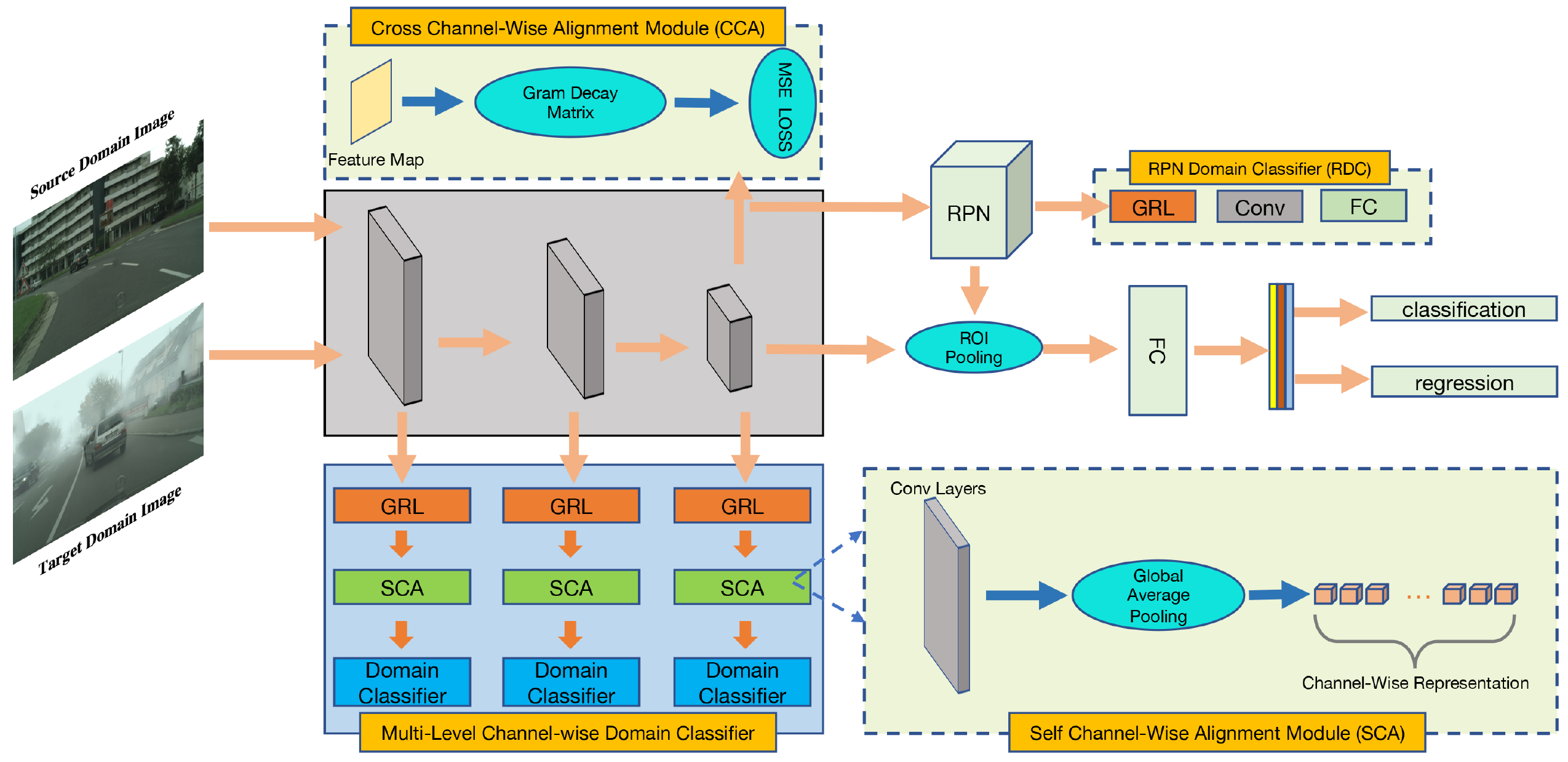} 
   \vspace{-2ex}
   \caption{The network structure of our method. Our proposed method consists of three modules, that are \textit{Self channel-wise alignment module}, \textit{Cross channel-wise alignment module} and \textit{RPN Domain Classifier}, donated as SCA, CCA and RDC respectively. The first component SCA is trying to reduce the domain gap through channel-wise representation alignment, which based on our reasonable assumption that each channel of feature map has its own attention region. The second component CCA pays more attention to aligning the highly semantic feature map through our proposed cross channel alignment method. The final component RDC is proposed to increase the robustness of RPN module through a domain classifier. It can improve the performance of whole model, due to more robust proposals can be generated by a domain-invariant RPN subnetwork. We use Faster-RCNN \cite{ren2015faster} as our detection model that consists of the RPN and head part. (\textbf{Best viewed in color.})}
   \label{fig:structure}
\end{figure*}

\begin{figure}[t]
   \centering
   \includegraphics[width=0.8\linewidth]{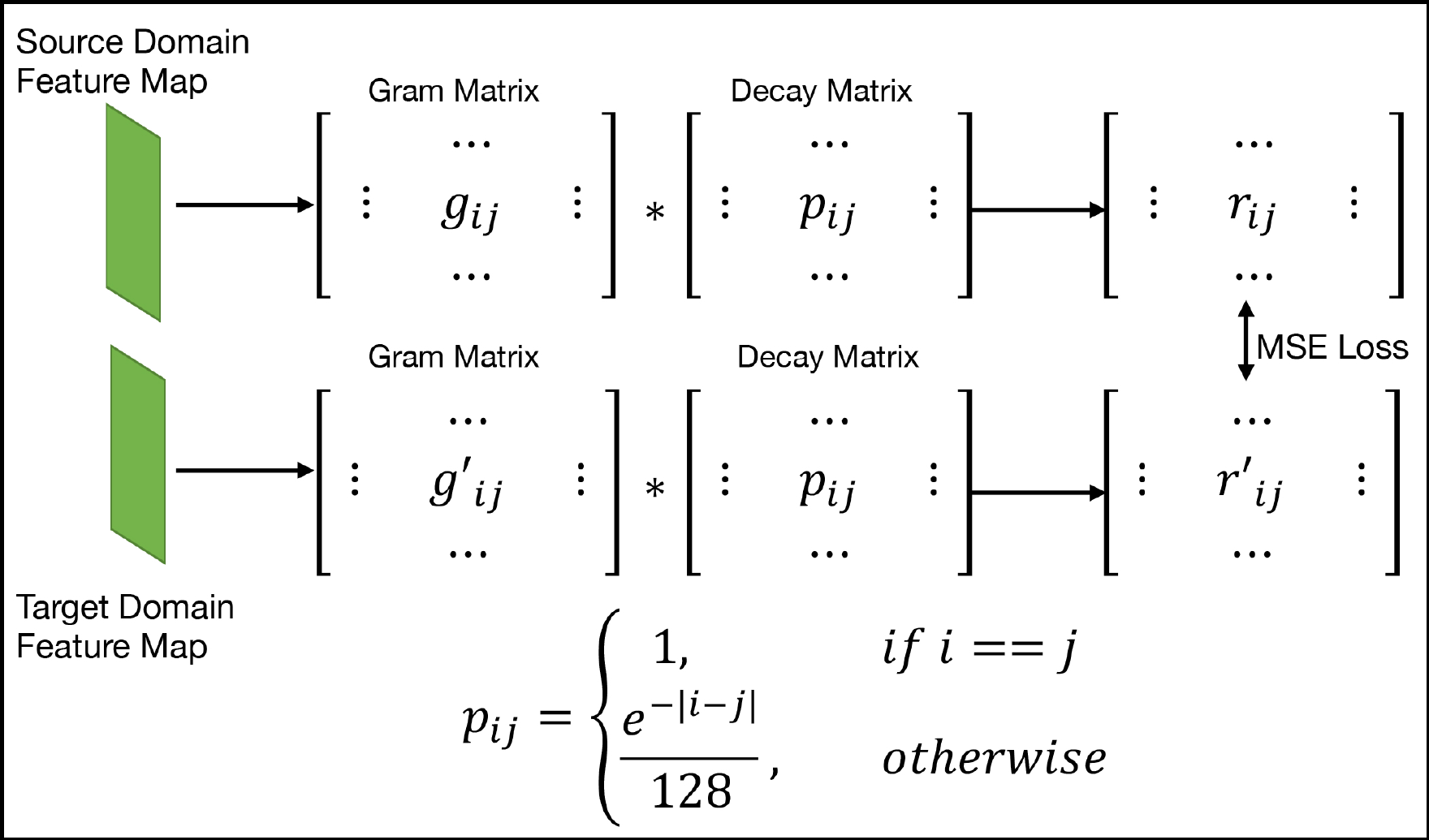} 
   \vspace{-2ex}
   \caption{Decay gram matrix loss workflow. We first obtain the gram matrix $G_{g}$ and $G_{g}^{'}$ on source domain and target domain. We denote decay matrix as $G_{d}$. Then we obtain $G_{dg}$ and $G_{dg}^{'}$ by multiplying decay matrix $G_{g}$ and $G_{g}^{'}$ with $G_{d}$. For the purpose of reducing the domain gap of high-level semantic representation between source domain and target domain, we conduct a \textit{mean square error} loss between the two decay gram matrix $G_{dg}$ and $G_{dg}^{'}$.}
   \label{fig:gram}
\end{figure}

\subsection{Framework Overview}
In the field of domain adaptation for detection, our target is to train a detector in one domain that can be adapted to another domain directly. To be more precise, we would like to learn the domain-invariant features by using the unsupervised method so that the learned model can perform well in both source and target domains. Inspired by the semantic coherence between channels and different objects in the image (as aforementioned), we propose two channel-wise alignment schemes and a new domain classifier on RPN that can drop into the Faster RCNN~\cite{ren2015faster} smoothly. These three additional modules can be regarded as the alignment losses to help optimize the model to a better status during end-to-end training. When inference, they will be removed so there is no additional computational cost for testing.

As illustrated in Fig.~\ref{fig:attention}, we observe that each channel is usually attentive to the particular semantic object (\textit{i.e.}, in each channel, the same class of objects will be prominent together), and different channels tend to focus on distinct objects, showing the diversity and informativeness across all channels.
These phenomena drive us to make a reasonable assumption that each channel on the feature map has its own regions of interest. Motivated by this, we explore the inner information within channels and the mutual relationship cross channels by proposing the self channel alignment and cross channel alignment modules. Furthermore, we also notice that the coarse regions generated by RPN contribute significantly in the Faster R-CNN \cite{ren2015faster} detector. Thus, we further draw attention on the RPN to obtain a \textit{domain-invariant} RPN module (by applying a domain classifier after RPN module) that can generate more robust coarse proposals.

\subsection{Channel-wise Alignment} \label{Channel-wise Alignment}
The channel-wise alignment contains two parts: self channel-wise alignment module and cross channel-wise alignment module. The former one is used to align the domain gap within channel interior and the latter can reduce or alleviate the domain gap between different channels.

\noindent \textbf{Self Channel-wise Alignment.} Self channel-wise alignment module $C_{s}$ aims at aligning the domain gap through channel interior information. As mentioned above, each channel has its own attentive region, thus we try to obtain the representation of each channel and then adopt the domain classifier to distinguish it whether from source domain or target domain. That is to say, the feature map extracted by backbone network will be fed into the self channel-wise alignment module for domain classification. This module has three parts: gradient reverse layer \cite{ganin2016domain}, convolution layers and average global pooling layer. The function of this three parts is to reverse the gradient of this module, further increase its nonlinear characteristics and obtain the representation of each channel individually. If the feature map we feed into this module has the shape of ($C$, $H$, $W$) where $C$ is the channel, $H$ is the height and $W$ is the width, we then can get a vector with the shape of ($C$, $1$). The objective $\mathcal L_{s}^r$ can be formulated as:
\begin{equation}
  \mathcal L_{s}^r = -\sum_{i, c}\left[D_{i}\log{p_{i}^{c}} + \left(1 - D_{i}\right)\log{\left(1 - p_{i}^{c}\right)}\right]. 
  \label{equ:self-channel-loss}
\end{equation}
where $r$ denotes the stage of backbone, e.g., for VGG16 \cite{vgg16}, the value of $r$ can be $\{1,2,3,4,5\}$. $\mathcal L_{s}$ is the objective loss of self channel-wise alignment.
$i$ is the index of $i$-th image during training. $D_{i}$ denotes domain label, \textit{i.e.}, $D_{i} = 0$ represents the source domain and $D_{i} = 1$ represents the target domain. $p_{i}^{c}$ denotes the probabilistic representation of the $c$-th channel.

We observe that it can achieve better integration if we apply the proposed module to all the stages of a network, as shown in Fig.~\ref{fig:structure}.
Thus, the total loss of this module can be formulated as:
\begin{equation}
    \mathcal L_{s} = \sum_{r} \mathcal L_{s}^r.
    \label{equ:sum-multi-level-channel-loss}
\end{equation}

Here we would like to simultaneously optimize the parameters of the domain classifier to minimize the above domain classification loss, and also optimize the parameters of the base network to maximize this loss. Thanks to the gradient reverse layer \cite{ganin2016domain}, we can train this module and the detection framework together through an end-to-end manner by inserting GRL to the beginning of each side branch.

\noindent \textbf{Cross Channel-wise Alignment. }
In order to further utilize the internal information of channels, we also explore the mutual relationship across different channels. Inspired by~\cite{styletransfer}, we find that gram matrix can be applied to represent the attribute/style of an images through high-level semantic representations. In our work, we utilize the gram matrix to evaluate the difference of attribute/style between source domain image and target domain images. Specifically, when given a feature map with the shape of ($C$, $H$, $W$), we first reshape it to ($C$, $H*W$), namely matrix $G_{ori}$. Each row in $G_{ori}$ can be regarded as the representation of each channel. The gram matrix $G_{g}$ is the matrix product between $G_{ori}$ and $G_{ori}^{T}$, \ie~ $G_{g} = G_{ori} \cdot G_{ori}^{T}$, where $G_{ori}^{T}$ is the transpose matrix of $G_{ori}$. Therefore, each element of $G_{g}$ can be considered as the mutual representation of current channel with another one. Besides, we further propose a decay matrix $G_{d}$ that is multiplied element-wisely by gram matrix $G_{g}$, based on the assumption that the mutual relation decreases as the distance between the channel to another channel increases. In some degree, our decay matrix can be seen as the special case of \cite{li2017demystifying} with a specific kernel, but note that our kernel is quite different with the kernel mentioned in \cite{li2017demystifying}, e.g. \textit{linear kernel, polynomial kernel and gaussian kernel}.  Our kernel, \textit{decay gram matrix} $G_{dg}$, is calculated with the following formulation:
\begin{equation}
    G_{dg} = G_{g} \cdot G_{d}.
    \label{equ:decay-gram-matrix}
\end{equation}
Note that $\cdot$ denotes the element-wise multiplication.
For any element \textit{p$_{ij}$} in
decay gram matrix \textit{G$_{dg}$}, we define it as the equation ~\ref{equ:decay-matrix}:
\begin{equation}
    p_{i,j} =
    \begin{cases}
        1 & \text{if}\ \ i=j, \\ 
        \frac{e^{-|i-j|}}{128} & \text{else}.
    \end{cases}
    \label{equ:decay-matrix}
\end{equation}

During the training, we can obtain two decay gram matrix, \textit{i.e.} $G_{dg}^{s}$ from the source domain and $G_{dg}^{t}$ from the target domain. We finally calculated the \textit{mean square error} (MSE) loss between this two decay gram matrix, which can be formulated by ~\ref{equ:cross-channel-wise-loss}:
\begin{equation}
    \mathcal L_{c} = \frac{1}{|N|} \sum_{i}\sum_{j} \| G_{dg_{(i,j)}}^{s} - G_{dg_{(i,j)}}^{t} \|^{2},
    \label{equ:cross-channel-wise-loss}
\end{equation}
where $N$ denotes the total number of elements of decay gram matrix $G_{dg}^{s}$ (or $G_{dg}^{t}$). 
The mechanism of cross channel-wise alignment module has been shown in Figure ~\ref{fig:structure} and Figure ~\ref{fig:gram}.

\subsection{RPN Domain Classifier}
\label{RPN Domain Classifier}
Previous works of domain adaptation on object detection try to obtain a domain-invariant backbone network mainly through aligning domain gap in image-level and instance-level \cite{chen2018domain} or local-strong and global-weak level \cite{saito2019strong}, or mining the relation of highly relative proposals \cite{zhu2019adapting}. Yet none of them pay attention to the robustness of RPN. We assume that if RPN module can be trained as much robust as possible, the proposals generated by RPN will be more domain-invariant, which can be fine-tuned later by Fast R-CNN \cite{ren2015faster} part. We also note that, RPN module has only two convolution layers, \textit{i.e.} classification convolution layer and regression convolution layer. These two layers can also play the role of sliding window as mentioned by \cite{ren2015faster}. Based on these findings, we do not use channel-wise alignment in RPN module, due to the function of RPN module does not obtain highly semantic feature, just generating proposals through sliding window mechanism instead. Thus, in this paper, we are trying to reduce the domain gap for the each active location on feature map that is fed into RPN. Besides, RPN generates the proposals exactly through every active location on feature map, so if we can reduce the domain gap of these active locations, we could improve the robustness of the whole model. 

Motivated by above, we first integrate gradient reverse layer \cite{ganin2016domain} into RPN module. Then we add three convolution layers to reduce the output channels by the factor 2 for the purpose of reducing the computational cost of the later FC layer. Finally, we get a one-dimensional vector through FC layer. We call these three parts as \textit{RPN domain classifier}. We use binary cross entropy loss as our optimize objective. The loss of RPN domain classifier denoted as equation~\ref{equ:rpn-domain-loss}:
\begin{equation}
    \mathcal L_{r} = -\sum_{i, u,v}\left[D_{i}\log{p_{(u,v)}^{i}} + \left(1 - D_{i}\right)\log{\left(1 - p_{(u,v)}^{i}\right)}\right].
    \label{equ:rpn-domain-loss}
\end{equation}
In equation~\ref{equ:rpn-domain-loss}, $i$ denotes the $i$-th image during training, $D_{i}$ denotes the domain label as mentioned above, $(u,v)$ denotes the active location on feature map which will be fed into domain classifier of RPN.

\subsection{Overall Objective}

As mentioned above, we have proposed three domain adaptive parts which are integrated into the Faster R-CNN, thus we divided the whole architecture into two parts: detection part \textit{i.e.} Faster R-CNN and domain adaptation part \textit{i.e.} \textit{self channel-wise alignment} module, \textit{cross channel-wise alignment} module and \textit{RPN domain classifier}.

Our total objective optimization $\mathcal L_{total}$ can be computed as the equation~\ref{equ:total-loss}:
\begin{equation}
    \mathcal L_{total} = \mathcal L_{det} + \lambda_{s}\mathcal L_{s} + \lambda_{c}\mathcal L_{c} + \lambda_{r}\mathcal L_{r},
    \label{equ:total-loss}
\end{equation}
where $\mathcal L_{det}$ is the Faster R-CNN loss function followed by \cite{ren2015faster}, as is shown in equation~\ref{equ:detection-loss}:
\begin{equation}
    \mathcal L_{det} = \mathcal L_{cls} + \mathcal L_{loc},
    \label{equ:detection-loss}
\end{equation}
where $\mathcal L_{cls}$ is the cross-entropy loss function and $\mathcal L_{loc}$ denotes the smooth L1 loss function. Note that $\mathcal L_{det}$ only works for source domain because of there is no label for target domain. $\lambda_{s}$, $\lambda_{c}$ and $\lambda_{c}$ are trade-off parameter, which are all set to be 1 in our experiments.

\subsection{Implementation Details}

\noindent \textbf{Self Channel-wise Alignment. }
In this part, for each stage, we first add GRL \cite{ganin2016domain} layer after the feature map that will be fed into the whole domain classifier. Then two convolution layers are added following the GRL layer, both kernel size are 3$\times$3, output channel that equals to input channel, stride is 1, and set padding to be 1. One BN \cite{bn} layer is added after the first convolution layer to reduce the training volatility. Finally we use global average pooling layer to obtain the representation of each channel-wise feature map. Note that we have 5 stages for VGG16 that all have self channel-wise alignment module.

\noindent \textbf{RPN Domain Classifier. }
In this part, we no longer to obtain the representation of channel-wise feature map, the reason is mentioned by Section ~\ref{RPN Domain Classifier}. Similar to \textit{self channel-wise alignment} module, we add a GRL \cite{ganin2016domain} layer firstly. Then three convolution layers are added, which has the kernel size of 3$\times$3, stride of 1 and padding of 1. The output channel is half of input channel for each convolution layer for reducing the complexity of computation, which will be used by FC layer later. After the third convolution layer, we add a FC layer to get one dimensional vector which represent each active location to generate proposals later.

\noindent \textbf{Training Details. }
We use Faster R-CNN \cite{ren2015faster} as our detection model followed by \cite{chen2018domain} with VGG16 \cite{vgg16} backbone network. The hyper-parameters of detection part follow \cite{ren2015faster}. The detection part is initialized using weights pre-trained on ImageNet \cite{imagenet}.

During training, we resize the short size of 600 pixels to fit in GPU memory, set the batch size to 1 (\textit{i.e.} one source image and one target image at every iteration). Note that, for KITTI \cite{kitti} dataset, we just remain the short size to 375 pixels, due to the image shape of KITTI \cite{kitti} is 375$\times$1250. We fine-tune the whole model with learning rate of 0.001 for 15 epochs and 0.0001 for later 10 epochs and weight decay of 5$\times$1e-4. Note that, for fair comparison, we report mean average precision (mAP) with a threshold of 0.5 for evaluation (\textit{i.e.} AP50) follows \cite{chen2018domain}.

%% file: exper.tex
\begin{table*}[t]
    \centering
    \vspace{-2ex}
    \begin{tabular}{c|cccccccc|c}
         \hline 
         Methods & person & rider & car & truck & bus & train & motorbike & bicycle & mAP \\
         \hline \hline
         Source-only  & 28.2 & 32.2 & 39.1 & 15.9 & 24.5 & 13.1 & 19.0 & 28.7 & 25.1 \\
         \hline
         DA-Faster \cite{chen2018domain} & 26.3 & 35.3 & 45.5 & 29.2 & 41.0 & 21.4 & 20.8 & 23.2 & 30.3 \\
         SCDA \cite{zhu2019adapting}  & 33.5 & 38.0 & 48.5 & 26.5 & 39.0 & 23.3 & 28.0 & 33.6 & 33.8 \\
         SWDA \cite{saito2019strong} & 29.9 & 42.3 & 43.5 & 24.5 & 36.2 & 32.6 & 30.0 & 35.3 & 34.3 \\
         \hline
         Ours  & \textbf{37.2} & \textbf{44.1} & \textbf{54.8} & \textbf{30.6} & \textbf{46.9} & \textbf{37.1} & \textbf{30.7} & \textbf{36.8} & \textbf{39.8} \\
         \hline
         Oracle & 39.5 & 46.9 & 57.9 & 34.2 & 53.7 & 31.9 & 35.7 & 38.8 & 42.3 \\
         \hline
    \end{tabular}
    \caption{Normal-to-foggy domain adaptation}
    \label{tab:normal-foggy}
\end{table*}

In this section, we first introduce our datasets in details. Then we provide comprehensive results of domain adaptation for object detection in different settings, including normal-to-foggy, cross-camera adaptation and synthetic-to-real. These settings are mainly referred from \cite{zhu2019adapting}. In the third part, we analyze our proposed method via two ablation studies. Finally, we conduct our experiment on the instance segmentation task to further verify the scalability and generality of our proposed method.

\subsection{Datasets and Evaluation}

We evaluate our proposed method on four datasets, including Cityscapes, Foggy-Cityscapes, KITTI and SIM-10k. We use these four datasets conduct four experiments. For normal-to-foggy domain adaptation, we experiment on Cityscapes and Foggy-Cityscapes. For cross-camera domain adaptation, we experiment on Cityscapes and KITTI. For synthetic-to-real domain adaptation, we experiment on Cityscapes and SIM-10k. For domain adaptation on instance segmentation task, we experiment on Cityscapes and Foggy-Cityscapes.

\subsection{Main Results on Domain Adaptive Detection}
The detection framework we used in our experiment is Faster-RCNN. We compare our method with current state-of-the-art works \cite{zhu2019adapting,chen2018domain,saito2019strong}. We observe that some other methods are either under weakly supervised or class-specific settings, which cannot be directly compared here.

\begin{table}[t]
    \centering
    \begin{tabular}{c|c}
        \hline
        Methods & AP on Car \\
        \hline \hline
        Source-only & 37.5 \\
        \hline
        DA-Faster \cite{chen2018domain} & 38.9 \\
        SCDA \cite{zhu2019adapting} & 43.0 \\
        SWDA \cite{saito2019strong} & 41.5 \\
        \hline
        Ours  & \textbf{43.2} \\
        \hline
        Oracle & 60.5 \\
        \hline
    \end{tabular}
    \caption{Synthetic-to-real domain adaptation}
    \label{tab:sim10k-cityscapes}
\end{table}
\begin{table}[t]
    \centering
    \begin{tabular}{c|c}
        \hline
        Methods & AP on Car \\
        \hline \hline
        Source-only & 36.1 \\
        \hline
        DA-Faster \cite{chen2018domain} & 38.5 \\
        SCDA \cite{zhu2019adapting} & 42.5 \\
        SWDA$^{*}$ \cite{saito2019strong} & 41.6 \\
        \hline
        Ours & \textbf{43.7} \\
        \hline
        Oracle & 48.6 \\
        \hline
    \end{tabular}
    \caption{Cross-camera domain adaptation}
    \label{tab:kitti-cityscapes}
\end{table}
\begin{table*}[t]
    \centering
    \begin{tabular}{c|ccc|cccccccc|c}
         \hline 
         Methods & SCA & CCA & RDC & person & rider & car & truck & bus & train & motorbike & bicycle & mAP \\
         \hline \hline
         Source-only &  &  &  & 28.2 & 32.2 & 39.1 & 15.9 & 24.5 & 13.1 & 19.0 & 28.7 & 25.1 \\
         \hline
         \multirow{7}*{Ours} & \checkmark &  &  & 37.2 & 43.1 & 53.9 & 28.2 & 47.1 & 30.8 & 30.4 & 37.1 & 38.5 \\
                 &  & \checkmark  &  & 28.1 & 33.8 & 39.7 & 16.1 & 25.2 & 15.0 & 17.1 & 29.5 & 25.6 \\
                 &  &  & \checkmark & 29.7 & 33.9 & 41.9 & 17.7 & 29.2 & 18.3 & 20.9 & 30.0 & 27.7 \\
                 & \checkmark & \checkmark & & 37.3 & 43.5 & 54.3 & 29.3 & \textbf{47.7} & 33.4 & \textbf{31.5} & \textbf{37.6} & 39.3 \\
                 & \checkmark & & \checkmark & \textbf{37.5} & \textbf{44.5} & 54.0 & 29.2 & 46.4 & 31.1 & 30.1 & \textbf{37.6} & 38.8 \\
                 & & \checkmark & \checkmark & 31.8 & 38.8 & 46.7 & 20.5 & 32.8 & 12.9 & 23.4 & 32.3 & 29.9 \\
                 & \checkmark & \checkmark  & \checkmark & 37.2 & 44.1 & \textbf{54.8} & \textbf{30.6} & 46.9 & \textbf{37.1} & 30.7 & 36.8 & \textbf{39.8} \\
         \hline
    \end{tabular}
    \caption{Ablation study from Cityscapes to Foggy-Cityscapes}
    \label{tab:ablation-study-foggy}
\end{table*}

\noindent \textbf{Normal to Foggy. }
In this experiment, we use Cityscapes  as our source domain dataset and Foggy-Cityscapes as our target dataset. We train the model on the source domain with the unlabelled target domain training images, and evaluate on the target domain testing set. We also compare our results on source-only results, which means the model are trained only on the source domain and tested on the target one. 

The results are illustrated in the Table~\ref{tab:normal-foggy}. From this table, we can see that our results remarkably outperform the existing state-of-the-art method \cite{saito2019strong} with a large margin, i.e., about 5.5\% improvement (39.8\% v.s. 34.3\%). 
This result demonstrates that our proposed method is more robust for object detection under different weather conditions.

\noindent \textbf{Cross Camera Adaptation. } In this experiment, we use KITTI dataset and Cityscapes dataset for training and evaluation. The purpose that we conduct this adaptation is in the real world scenarios, different camera sets always exist commonly like in autonomous driving so it's necessary to verify on this setting. KITTI consists of images with the size of 1250 $\times$ 375, while the size of images in Cityscapes  is 2048 $\times$ 1024. They are totally different in the filed-of-view. 
We train our model by cropping KITTI  to 2:1 with shorter edge as 375. Then we test on Cityscapes  also with the shorter edge as 2:1. The results are illustrated in Table ~\ref{tab:kitti-cityscapes}. From this table, we can observe that our results are higher than SCDA. We get about 1\% improvement.

\noindent \textbf{Synthetic to Real. }In this part, we introduce our experiments on different generation of domain adaptation. We take synthetic-to-real as an example. Due to the huge cost of collection and annotation of training data, people are more focusing on the usage of synthetic data. However, if we use the synthetic data to train the model directly, the accuracy will have a significant drop on the real data. Thus, improvement on synthetic-to-real scenario is desired urgently. 

In our experiment, we use SIM-10k as synthetic data and Cityscapes  as real data. We only train the model on the \textsl{Car} category because of the limitation of provided bounding boxes. We first train our model on the labeled SIM-10k  and label-free Cityscapes . Then we test on \textsl{car} class on Cityscapes  dataset. The results are illustrated on Table~\ref{tab:sim10k-cityscapes}. From the table, we can see that our result is better than SCDA \cite{zhu2019adapting}.



\begin{table*}[t]
    \centering
    \begin{tabular}{c|ccccc|cccccccc|c}
         \hline
         Method & S5 & S4 & S3 & S2 & S1 & person & rider & car & truck & bus & train & motorbike & bicycle & mAP \\
         \hline \hline
         \multirow{5}*{Ours} & \checkmark & & & & & 33.5 & 39.6 & 49.3 & 23.0 & 37.6 & 17.0 & 24.3 & 32.9 & 32.2 \\
                & \checkmark & \checkmark & & & & 35.4 & 41.2 & 51.5 & 24.4 & 43.5 & 22.8 & 27.7 & 34.1 & 35.1 \\
                & \checkmark &  & \checkmark & & & 36.3 & 41.6 & 52.0 & 25.5 & 45.0 & 24.8 & 29.4 & 34.2 & 36.4\\
                & \checkmark & \checkmark & \checkmark & & & 36.8 & 42.4 & 52.9 & 26.5 & 45.1 & 31.5 & 30.7 & 35.8 & 37.7 \\
                & \checkmark & \checkmark & \checkmark & \checkmark & & 37.2 & 43.8 & 53.8 & 27.1 & \textbf{48.9} & 32.2 & 28.4 & 36.6 & 38.5 \\
                & \checkmark & \checkmark & \checkmark & \checkmark & \checkmark & \textbf{37.2} & \textbf{44.1} & \textbf{54.8} & \textbf{30.6} & 46.9 & \textbf{37.1} & \textbf{30.7} & \textbf{36.8} & \textbf{39.8} \\
         \hline
    \end{tabular}
     \caption{Ablation study of different combination of SCA. S indicates stage, which means different layer of our backbone network VGG16 that we add SCA. For example, S5 indicates the fifth stage of VGG16, which is the highest level representation }
    \label{tab:SCA_combination}
\end{table*}

\subsection{Ablation Studies}
To give an in-depth analysis of our model, we conduct extensive ablation studies on normal-to-foggy setting. We first analyse the improvement of our all proposed modules. Then we analyse the improvement of different combination of our self channel-wise alignment module.

\noindent \textbf{Different Modules Analysis. }
As illustrated in Table \ref{tab:ablation-study-foggy}, we first investige the effectiveness of the main components, including SCA, CCA and RDC respectively. SCA is Self Channel-Wise Alignment Module. CCA is Cross Channel-Wise Module and RDC is RPN Domain Classifier.

We first verify SCA, CCA and RDC independently, and find SCA has the highest improvement than the other two. It yields an improvement of 13.5\% over the Source-only method. It indicates that our proposed self channel-wise method has the highest contribution. The CCA module also can improve 0.6\% than the Source-only method. The RDC can improve 2.7\%. Then we compose these three modules one with another and find SCA with CCA can reach to 39.3\%, which is highest than SCA with RDC and CCA with RDC. It indicates that our sef channel-wise method and cross channel-wise method can work complementarily. It proves our assumption that the channel information can improve the total accuracy largely. Finally, we compose all the three modules and obtain the highest results. Besides, we also conduct the extensive experiments to demonstrate the effectiveness  of our decay matrix, as mentioned in Section \ref{Channel-wise Alignment}. The results show that our decay matrix indeed improve performance on all tasks, as is shown in Tabel \ref{tab:decay matrix}.

\begin{table}[t]
    \centering
    \begin{tabular}{c|c|c|c}
        \hline
        Methods & cs2foggy & sim10k2cs & kitti2cs \\
        \hline \hline
        Source-only & 25.1 & 37.5 & 36.1 \\
        \hline
        without decay matrix & 25.3 & 37.9 & 36.4 \\
        decay matrix & \textbf{25.6} & \textbf{38.5} & \textbf{36.5} \\
        \hline
    \end{tabular}
     \caption{Extensive study to demonstrate the effectiveness of our decay matrix. Cs2foggy, sim10k2cs, kitti2cs means the source domain is Cityscapes, Sim10k and Kitti respectively. And the target domain is Foggy-Cityscapes and Cityscapes  respectively}
    \label{tab:decay matrix}
\end{table}

\noindent \textbf{Analysis of Different Combination of SCA. }
For analyzing the influence of different combination of SCA, we conduct experiment by adding SCA on different layers of backbone network VGG16. 
The results are demonstrated on Tab.~\ref{tab:SCA_combination}. There are 5 stages in VGG16, which can be concluded as conv1-*, conv2-*, conv3-*, conv4-* and conv5-*. We add SCA on every stage after pooling layer except for conv5-*. From Tab.~\ref{tab:SCA_combination}, we can find that with the increasing of stage, the mAP become higher. After adding SCA on all stages, we get the highest results.

\begin{table}[t]
    \centering
    \begin{tabular}{c|c|c}
        \hline
        Methods & mAP(Box) & mAP(Mask) \\
        \hline \hline
        Faster R-CNN & 25.0 & - \\
        \hline
        Source-only & 25.6 & 19.0 \\
        \hline
        SCDA \cite{zhu2019adapting} & 38.4 & \textbf{31.4} \\
        \hline
        Ours & \textbf{39.4} & 31.1 \\
        \hline
    \end{tabular}
    \caption{Domain adaptation for instance segmentation. We conduct this experiment from Cityscapes to Foggy-Cityscapes. Faster R-CNN is only trained on source domain with bounding boxes, thus it only have mAP with bounding boxes. We train our proposed method model on dataset with both bounding boxes and mask}
    \label{tab:instance-seg}
\end{table}


\subsection{Domain Adaptation for Instance Segmentation }
In this section, we introduce our further research on domain adaptation for instance segmentation. Instance segmentation task not only has the annotated bounding boxes, but also has the labelled mask, it can be used for verifying the scalability of our proposed method.

We mainly use our SCA and CCA in Mask R-CNN for this experiment. The network's backbone is also VGG16. The datasets we use for this experiment are Cityscapes  and Foggy-Cityscapes . The reason why we use this two datasets for this experiment is that they both have annotated mask and bounding boxes. The domain adaptation is from Cityscapes  to Foggy-Cityscapes . We train our model on the labelled Cityscapes  and label-free Foggy-Cityscapes. Then we evaluate the model on labelled Foggy-Cityscapes. The detection threshold that we set for calculating mAP is 0.7.

The results are reported in Table~\ref{tab:instance-seg}. First, comparing Faster R-CNN with Source-only, we can observe that it improves about 0.6\% by adding the mask branch. Comparing our results with Source-only, we can see our proposed method yields significant improvement, where the improvements are 3.8\% on mAP (Box) and 12.1\% on mAP (Mask). In addition, we also compare our results with current state-of-the-art method SCDA, it also improves about 1\% on mAP (Box) and comparable mAP (Mask). Furthermore, we visualize the domain adaptation results and compare them with Source-only results in the supplementary materials.

%% file: conclusion.tex
In this paper, we propose a channel-wise domain adaptation method for object detection. The channel-wise method consists of self channel-wise alignment(SCA) and cross channel-wise alignment(CCA) module. These two modules have remarkable improvement on the accuracy of detection from source domain to target domain. In addition, we also add RPN domain classifier(RDC) for further improvement. Furthermore, we experiment our proposed method on instance segmentation. The results outperform over Source-only and are higher than current state-of-the-art method. It demonstrates that our method has good scalability on region-based domain adaptation field.